\documentclass[letterpaper]{article}
\usepackage{aaai}
\usepackage{times}
\usepackage{helvet}
\usepackage{courier}
\usepackage[numbers]{natbib}
\usepackage{graphicx}
\usepackage{subfig}
\usepackage{mathrsfs}
\nocopyright
\begin{document}
%
\title{Multi-Target, Multi-Camera Tracking by Hierarchical Clustering: \\Recent Progress on DukeMTMC Project}
\author{Zhimeng Zhang, Jianan Wu, Xuan Zhang, Chi Zhang \\ Megvii Inc. \\ {zhangzhimeng, wjn, zhangxuan, zhangchi}@megvii.com}

\maketitle
\begin{abstract}
\begin{quote}
Although many methods perform well in single camera tracking, multi-camera tracking remains a challenging problem with less attention. DukeMTMC is a large-scale, well-annotated multi-camera tracking benchmark which makes great progress in this field. This report is dedicated to briefly introduce our method on DukeMTMC and show that simple hierarchical clustering with well-trained person re-identification features can get good results on this dataset.
\end{quote}
\end{abstract}

\section{Introduction}
Multi-target, multi-camera tracking is an useful but challenging problem in computer vision. By combining the trajectory information from multiple cameras, it overcomes the limitation of the field of view in single camera and allows people to conduct a more detailed analysis of the target. This technology can be used in video surveillance, sports analysis and many other fields. However, compared to single camera tracking, multi-camera tracking has less location information and less strict time constraints, which makes it even more difficult. Some algorithms have been proposed to solve this knotty problem. E Ristani et al. \cite{ristani2016MTMC} formulated the tracking problem as correlation clustering problem. YT Tesfaye et al. \cite{tesfaye2017multi} formulated the tracking problem as finding constrained dominant sets from a graph. A Maksai et al. \cite{maksai2017non} put the tracking problem under global, non-Markovian behavioral constraints. In this report, we study a strategy based on the hierarchical clustering and show that it works well with person re-identification features.

\section{DukeMTMC Dataset}
DukeMTMC \cite{ristani2016MTMC} dataset is a multi-target, multi-camera tracking dataset. It consists of 85 minutes of 1080p, 60 fps videos recorded by 8 calibrated and synchronized static outdoor cameras, together with manual annotations as ground truth. This dataset is split into one training/validation set and two test sets, the easy test set and the hard test set. The training/validation set is 50 minutes long and the test sets totalled 35 minutes long. More than 7,000 single camera trajectories and over 2,000 identities appeared in the dataset in total, which is a very large number compared to other existing MTMC datasets, allowing to better evaluate the performance of tracking methods.

\section{Proposed Method}
In this section we show the details of our method. It includes four main steps: detection, feature extraction, single camera tracking and multi-camera tracking.
\begin{itemize}
    \item Detection: we studied both public detections and private detections. We use a Faster R-CNN \cite{Girshick2015FastR} detector to generate private detections.
    \item Feature Extraction: we use a person re-identification model to extract the appearance features.
    \item Single camera tracking: we merge neighbor bounding boxes into tracklets (short but reliable trajectories) first, then generate trajectories by hierarchically clustering tracklets.
    \item Multi-camera tracking: we merge the trajectories by a hierarchical clustering without distance matrix updating.
\end{itemize}

\subsection{Detection}
Private person detections are generated using the Faster R-CNN object detector, which works in two stages: First, it extracts candidate regions of interest (RoI) using a Region Proposal Network (RPN); Then it performs binary classification into person and background after applying ROI pooling on the candidate regions. We trained our detection model on COCO \cite{Lin2014MicrosoftCC} dataset and some other private datasets. Note that our detection model have not fine-tuned on DukeMTMC dataset though it can definitely improve the overall performance. We set a threshold value of 0.9 and perform non-maximum suppression (NMS) for each frame with iou threshold 0.3. Our detections can be download at https://pan.baidu.com/s/1hrPxYNq. 8 text files are provided. Each one contains detections in each camera with the following line format (each line represents a bounding box): [camera, frame, left, top, right, bottom, confidence].

For public detections, we set a threshold value of 0 and the nms threshold as 0.3.

\subsection{Person Re-Identification}
In order to combine the bounding boxes together with their appearance, we trained a person re-identification network to give an appearance feature for each bounding box. We trained our network following the AlignedReID \cite{zhang2017alignedreid} which gives an impressive result on reid task recently. Our model is trained on 7 public reid datasets, including CUHK03 \cite{Li2014DeepReIDDF}, CUHK-SYSU \cite{Xiao2017JointDA}, Market1501 \cite{Zheng2015ScalablePR} ,MARS \cite{Zheng2016MARSAV},  VIPeR \cite{Gray2008ViewpointIP}, PRID \cite{Hirzer2011PersonRB}, as well as DukeMTMC-reID \cite{zheng2017unlabeled} which is an extension of DukeMTMC.

In practice, we found that the reid model with higher top1 accuracy does not always lead to a better multi-camera tracking result.
Hence, we trained a smaller and faster reid model, whose top1 accuracy is slightly lower, but performs as well as the big one for multi-camera tracking.

Table \ref{tab:reid} shows the top 1 accuracy of our two reid models on the datasets, without re-ranking.

\begin{table}[]
    \centering
    \begin{tabular}{|c|c|c|}
    \hline Benchmark & big model & small model\\
    \hline
    \hline CUHK03    & 92.5 & 88.1\\
    \hline CUHK-SYSU & 96.0 & 92.1 \\
    \hline DukeMTMC-reID & 81.9 & 78.9\\
    \hline Market1501& 93.9 & 91.8\\
    \hline MARS      & 81.9 & 83.5 \\
    \hline
    \end{tabular}
    \caption{Top1 accuracy of our reid models without re-ranking}
    \label{tab:reid}
\end{table}

\subsection{Single Camera Tracking}
Single camera tracking is a popular research topic in computer vision and a lot of algorithms have been proposed \cite{2017:Leal:arxiv}. In this work, we proposed a simple method to aggregate bounding boxes into some trajectories in single camera and these trajectories will be further merged by our multi-camera tracking algorithm. Our single camera tracking strategy is a near online \cite{Choi2015NearOnlineMT} algorithm. We first merge bounding boxes of adjacent frames into tracklets by the Kuhn-Munkras algorithm \cite{Kuhn2010TheHM} in a small window. Then we use a hierarchical clustering to merge the tracklets into trajectories.

\subsubsection{Merging Neighbors}
Let $F_t$ and $F_{t+1}$ be two adjacent frames, where $t_{start} \leq t \leq t_{end}$ and $t_{start}$, $t_{end}$ are the range of the tracking window. Let $\{B^t_i\}_{i=1...N_t}$ be the bounding boxes in $F_t$, where ${N_t}$ is the number of bounding boxes in $F_t$. For each possible assignment between $\{B^t_i\}_{i=1...N_t}$ and $\{B^{t+1}_j\}_{j=1...N_{t+1}}$, we define a weight $W^t_{ij}$ as the likelihood that the two bounding boxes are the same person.
$W^t_{ij}$ is decided by two factors: the location closeness and the appearance similarity.
For location closeness, we force the IoU of the bounding boxes greater than a threshold, which is set to $0.5$ in this work.
For appearance similarity, we apply the L2 distance of their reid features.
Hence, $W^t_{ij}$ is computed by the following equation:
\begin{equation}
\label{eq:weight}
W^t_{ij} = s - || f^t_i - f^{t+1}_j ||_2 - g(IoU(B^t_i, B^{t+1}_j), 0.5),
\end{equation}
where $s$ is a given threshold, $f^t_i$ and $f^{t+1}_j$ are the reid features of $B^t_i$ and $B^{t+1}_j$, $IoU$ is their intersection over union, and $g$ is defined as
\begin{equation}
\label{eq:g}
g(x, k) =\left\{
\begin{array}{ll}
\infty, & x<k, \\
0, & x\ge k,
\end{array}\right.
\end{equation}
Then it can be solved as a maximum weighted bipartite matching problem by the Kuhn-Munkras algorithm.
Bounding boxes that are not matched are considered as false alarms and discarded.

\subsubsection{Hierarchical Clustering}
Hierarchical clustering is used to merge trajectories and tracklets together. We first calculate the distance matrix between them.
The distance between a trajectory and a tracklet is comprised of three parts:
The distance for appearance similarity, the distance of the separation part and the distance of the overlapping part.

For single camera tracking, the reid features of the same person are quite close to each other in nearby frames.
Hence the distance of appearance similarity is decided by the neighboring frames around the endpoints of trajectories.
In more details, given two trajectories $T_i$ and $T_j$, assuming the last frame of $T_i$, which is noted as $t_i$, is before the last frame of $T_j$,
then find the first frame after $t_i$ in the trajectory $T_j$, denoted as $t_u$.
The reid features of $T_i$ and $T_j$ are calculated as the average of reid features in neighboring frames around $t_i$ and $t_u$ separately.
The distance of appearance similarity is the L2 distance of the reid features of $T_i$ and $T_j$.
For separation part, the speed is caculated by the bounding boxes in frame $t_i$ of trajectory $T_i$ and frame $t_u$ of trajectory $T_j$, where $t_i$ and $t_u$ have the same definition as before.
If the speed is less than a threshold, the distance is set to zero; otherwise it is set to $\infty$.
For overlapping part, the average IoU is calculated. The distance is set to zero or $\infty$, depending on whether the average IoU is greater than a threshold.

Then, as clustering progresses, trajectories are merged and the distance matrix updated until the minimum distance is greater than a threshold. In general, this threshold can be the same as $s$ in equation \ref{eq:weight}. After merged, trajectory interpolation and smoothing are performed but new reid feature of interpolated bounding boxes will not be extracted for the consideration of speed.

\subsection{Multi-Camera Tracking}
Our multi-camera tracking is a greedy procedure or a hierarchical clustering without distance matrix updating when clustering. We put all the trajectories in all the cameras together, for each trajectory, we average the reid features to get a more robust one. The distance matrix is defined as the Euclidean distance between the averaged features. Then we perform re-ranking \cite{Zhong2017RerankingPR} on the distance matrix. Re-ranking is a critical step to improve the accuracy of person re-identification. In our experiments, re-ranking can improve IDF1 of multi-camera tracking on training/validation set by about 2\%-3\%. After re-ranking, we merge the trajectories according to the distance from small to large until minimum distance reaches a threshold. In addition, in order to make tracking more accuracy, some restrictions are added to the merging procedure:
\begin{itemize}
    \item Two trajectories should be separated less than one minute. One minute is a statistic on the training/validation set.
    \item Person can not appear in different cameras at the same time, except for 3 pairs of cameras with overlapping areas, which are camera 2 and camera 8, camera 3 and camera 5 and camera 5 and camera 7.
\end{itemize}
Note in multi-camera tracking, we also do in-camera trajectories merging. We found it is very helpful for long-term tracking since the averaged reid features are more robust.

\section{Results}
We show benchmark results for our method and compare them with prior ones in this section. The measures and results are given by the benchmark available on motchallenge.net.

Table \ref{tab:easy_multi_pub}, Table \ref{tab:easy_multi_pri}, Table \ref{tab:hard_multi_pub} and Table \ref{tab:hard_multi_pri} compared the trackers performance on easy test set and hard test set, where MTMC\_ReIDp is our method on public detections and MTMC\_ReID is our method on private detections. We also compared to some anonymous methods. It is shown that, our method has achieved the state of the art in all the test settings.
\begin{table}[htb!]
    \centering
    \begin{tabular}{|c|c|c|c|}
    \hline Tracker & IDF1$\uparrow$ & IDP$\uparrow$  & IDR$\uparrow$\\
    \hline
    \hline MTMC\_ReIDp &\textbf{74.4}&\textbf{84.4}&\textbf{66.4}\\
    \hline MYTRACKER&65.4&71.1&60.6\\
    \hline MTMC\_CDSC \cite{tesfaye2017multi} &60.0&68.3&53.5\\
    \hline W. Liu et al. \cite{liu2017multi} & 55.5 &78.89&44.6\\
    \hline lx\_b   &58.0&72.6&48.2\\
    \hline BIPCC \cite{Ristani2016PerformanceMA}&56.2&67.0&48.4\\
    \hline dirBIPCC  &52.1&62.0&45.0\\
    \hline PT\_BIPCC \cite{maksai2017non} &34.9&41.6&30.1\\
    \hline
    \end{tabular}
    \caption{Multi-Camera Tracking Results of Easy Test Set on Public Detections}
    \label{tab:easy_multi_pub}
\end{table}

\begin{table}[htb!]
    \centering
    \begin{tabular}{|c|c|c|c|}
    \hline Tracker & IDF1$\uparrow$ & IDP$\uparrow$  & IDR$\uparrow$\\
    \hline
    \hline MTMC\_ReID  &\textbf{83.2}&\textbf{85.2}&\textbf{81.2}\\
    \hline DeepCC  &82.0&84.4&79.8\\
    \hline
    \end{tabular}
    \caption{Multi-Camera Tracking Results of Easy Test Set on Private Detections}
    \label{tab:easy_multi_pri}
\end{table}

\begin{table}[htb!]
    \centering
    \begin{tabular}{|c|c|c|c|}
    \hline Tracker & IDF1$\uparrow$ & IDP$\uparrow$  & IDR$\uparrow$\\
    \hline
    \hline MTMC\_ReIDp&\textbf{65.6}&\textbf{78.1}&\textbf{56.5}\\
    \hline MTMC\_CDSC \cite{tesfaye2017multi}&50.9&63.2&42.6\\
    \hline MYTRACKER &50.1&58.3&43.9\\
    \hline lx\_b     &48.3&60.6&40.2\\
    \hline BIPCC \cite{Ristani2016PerformanceMA}    &47.3&59.6&39.2\\
    \hline dirBIPCC &45.0&56.3&37.5\\
    \hline PT\_BIPCC \cite{maksai2017non} &32.9&41.3&27.3\\
    \hline
    \end{tabular}
    \caption{Multi-Camera Tracking Results of Hard Test Set on Public Detections}
    \label{tab:hard_multi_pub}
\end{table}

\begin{table}[htb!]
    \centering
    \begin{tabular}{|c|c|c|c|}
    \hline Tracker & IDF1$\uparrow$ & IDP$\uparrow$  & IDR$\uparrow$\\
    \hline
    \hline MTMC\_ReID&\textbf{74.0}&\textbf{81.4}&\textbf{67.8}\\
    \hline DeepCC    &68.5&75.9&62.4\\
    \hline
    \end{tabular}
    \caption{Multi-Camera Tracking Results of Hard Test Set on Private Detections}
    \label{tab:hard_multi_pri}
\end{table}

The hard test set and the easy test set are split according to the scale and the crowd, hence in the hard test set, location information will be more unreliable. In such a condition, more reliable reid feature can help a lot on tracking, which is the reason why we have a larger advantage on the hard test set.

\begin{figure*}[htb!]
\centering
\subfloat[]{
\includegraphics[width=3.5in]{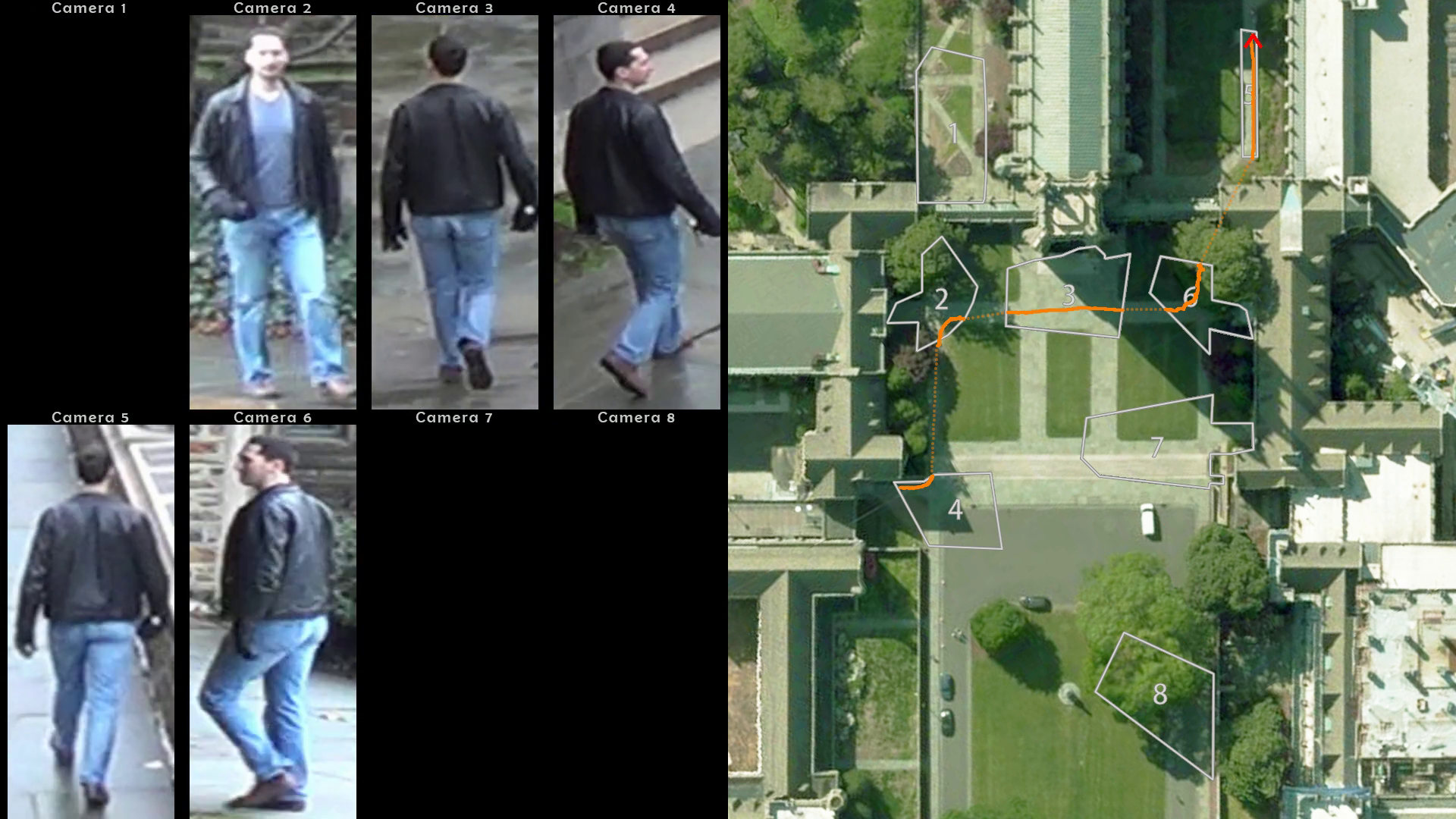}} 
\subfloat[]{
\includegraphics[width=3.5in]{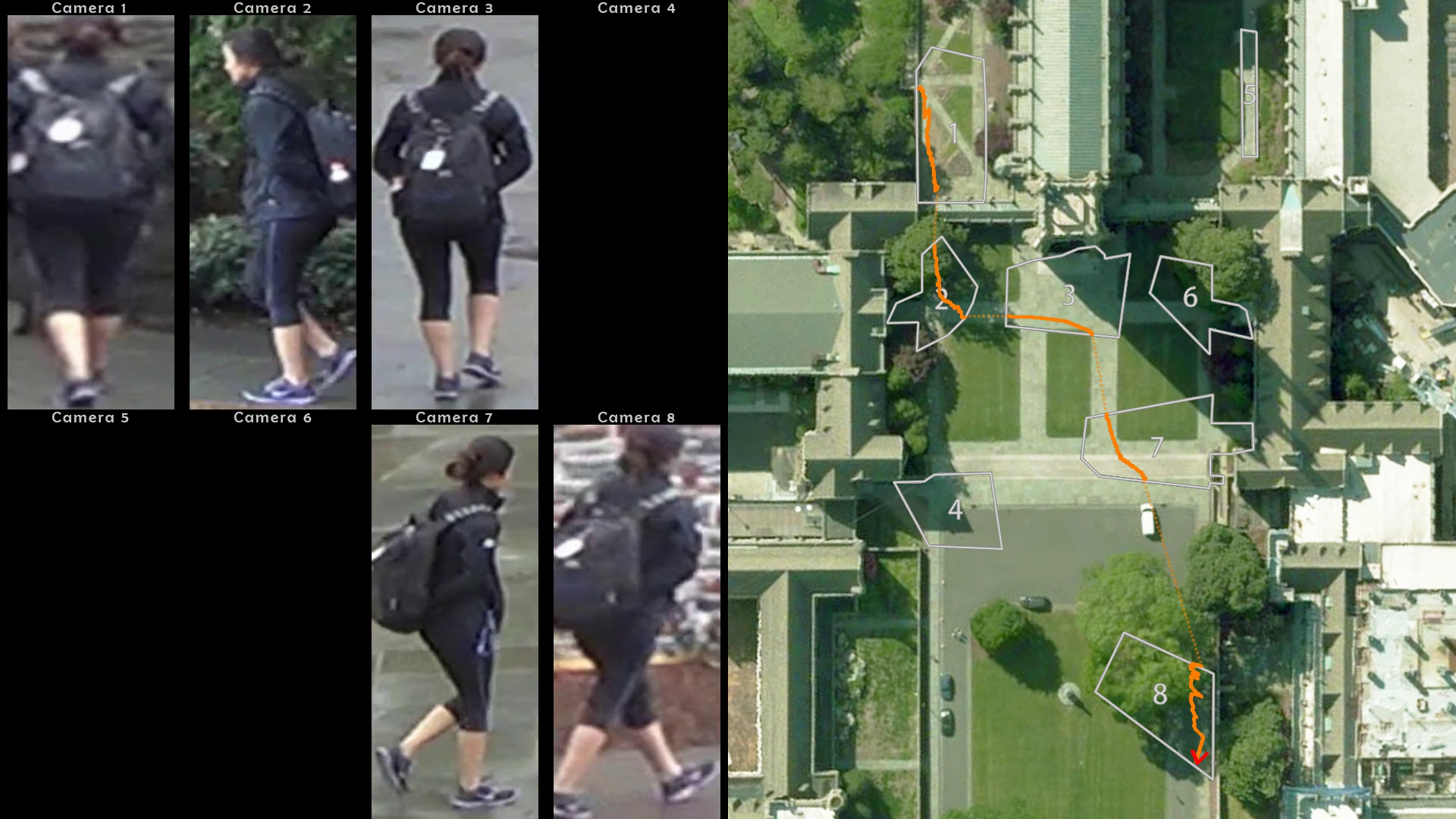}}\\
\subfloat[]{
\includegraphics[width=0.5\linewidth]{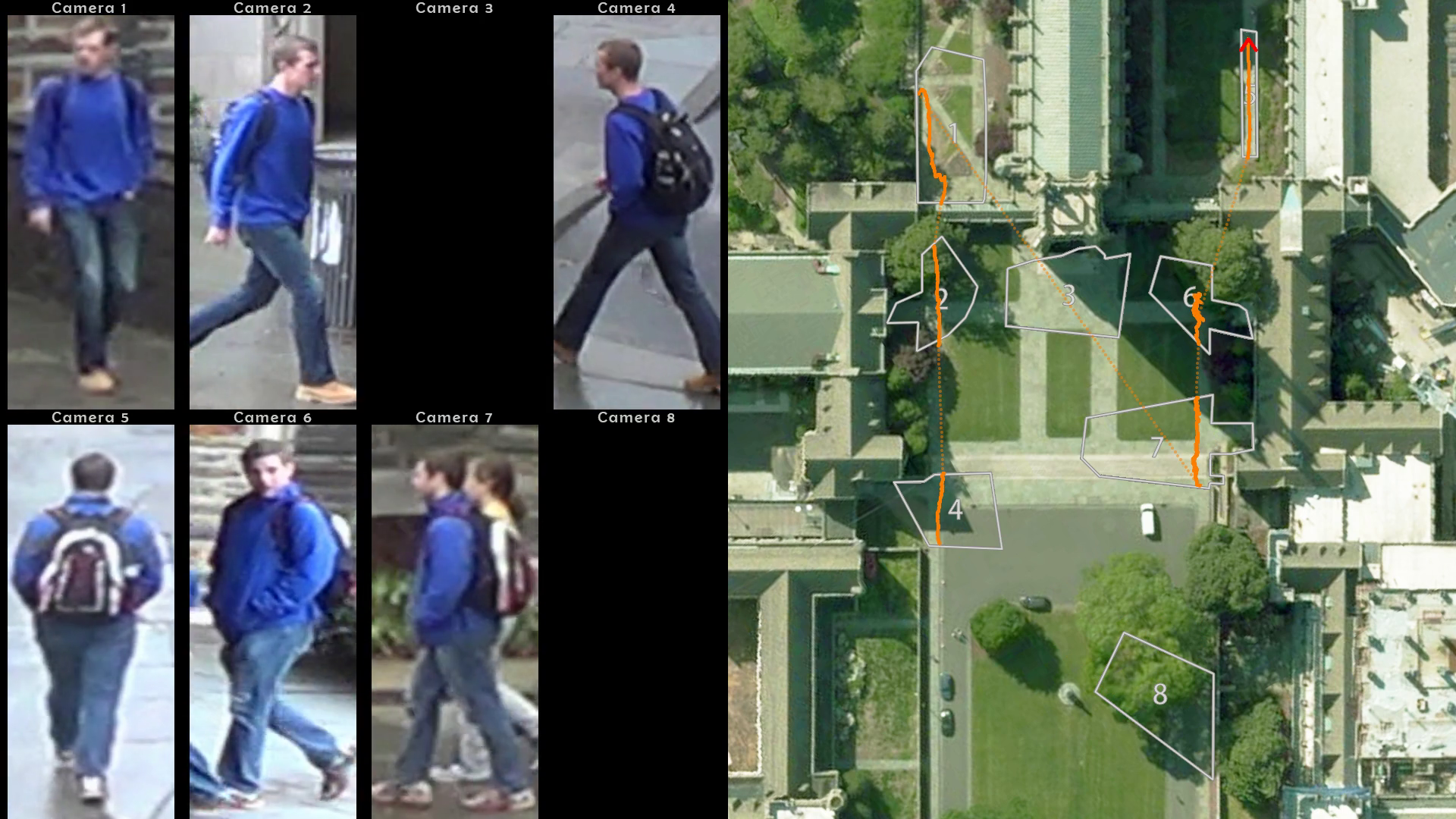}}
\subfloat[]{
\includegraphics[width=0.5\linewidth]{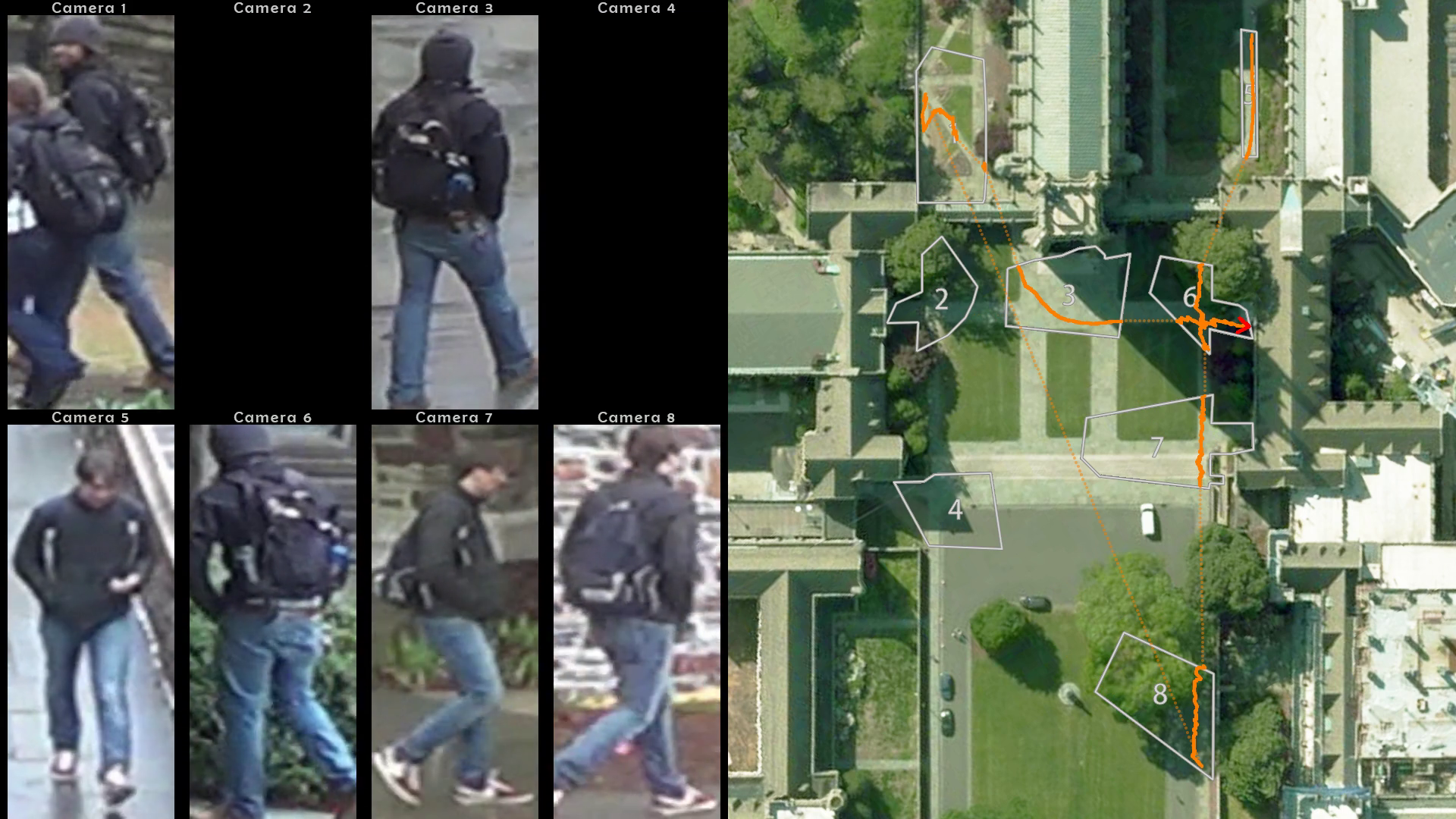}}\\
\caption{Examples of our results. (a), (b) are successful examples and (c), (d) are failure examples. The curves in the right are the trajectories and the dotted parts represent interpolated trajectories between the cameras. For better effect, in all the examples, cameras 1-8 have been renumbered to cameras 8, 7, 6, 5, 3, 1, 2, 4.}
\label{fig:examples}
\end{figure*}

Figure \ref{fig:examples} shows some examples of out results. According to our observation, most of the failures in tracking came from different people with similar appearances.
Some cases may be even difficult for human to distinct.

Note that our method is not an online algorithm, especially the multi-camera tracking part, but we can easily transform it to a near online one like single-camera tracking part does. We believe this will not lose too much accuracy.



\section{Conclusion}
In this report, we have presented our method for multi-target, multi-camera tracking on DukeMTMC dataset. We use hierarchical clustering algorithm to merge trajectories. The distance matrix is mainly based on high quality reid features. Benchmark results on easy test set and hard set set show that our method achieves new state of the art results. We hope this report can help the development of multi-camera tracking.

\bibliography{mtmc.bib}
\bibliographystyle{aaai} 
\end{document}